\documentclass[AMA,STIX2COL]{MRM}

\articletype{Research Article}%

\received{26 April 2016}
\revised{6 June 2016}
\accepted{6 June 2016}
\begin{document}

\title{HARP: HARmonizing in-vivo diffusion MRI using Phantom-only training}

\author[1]{Hwihun Jeong}{\orcid{0000-0002-5711-4290}}
\author[1,2]{Qiang Liu}{\orcid{0009-0007-6818-0115}}
\author[3]{Kathryn E. Keenan}{\orcid{0000-0001-9070-5255}}
\author[4,5]{Elisabeth A. Wilde}{}
\author[6]{Walter Schneider}{}
\author[6]{Sudhir Pathak}{\orcid{0000-0001-6669-2263}}
\author[7]{Anthony Zuccolotto}{\orcid{0009-0006-9067-4923}}
\author[8,9]{Lauren J. O’Donnell}{\orcid{0000-0003-0197-7801}}
\author[1]{Lipeng Ning}{\orcid{0000-0003-4992-459X}}
\author[1]{Yogesh Rathi}{\orcid{0000-0002-9946-2314}}

\authormark{Hwihun Jeong \textsc{et al}}

\address[1]{\orgdiv{Department of Psychiatry}, \orgname{Brigham and Women's Hospital, Harvard Medical School}, \orgaddress{\state{Boston, Massachusetts}, \country{United States of America}}}
\address[2]{\orgdiv{College of Engineering}, \orgname{Northeastern University}, \orgaddress{\state{Boston, Massachusetts}, \country{United States of America}}}
\address[3]{\orgdiv{Physical Measurement Laboratory},\orgname{National Institute of Standards and Technology}, \orgaddress{\state{Boulder, Colorado}, \country{United States of America}}}
\address[4]{\orgdiv{Department of Neurology}, \orgname{University of Utah School of Medicine}, \orgaddress{\state{Salt Lake City, Utah}, \country{United States of America}}}
\address[5]{\orgname{George E. Wahlen Veterans Affairs Medical Center}, \orgaddress{\state{Salt Lake City, Utah}, \country{United States of America}}}
\address[6]{\orgname{University of Pittsburgh}, \orgaddress{\state{Pittsburgh, Pennsylvania }, \country{United States of America}}}
\address[7]{\orgname{Psychology Software Tools}, \orgaddress{\state{Pittsburgh, Pennsylvania }, \country{United States of America}}}
\address[8]{\orgdiv{Department of Radiology}, \orgname{Brigham and Women's Hospital, Harvard Medical School}, \orgaddress{\state{Boston, Massachusetts}, \country{United States of America}}}
\address[9]{\orgname{Harvard-MIT Health Sciences and Technology}, \orgaddress{\state{Cambridge, Massachusetts}, \country{United States of America}}}

\abstract[Abstract]{
\section{Purpose} Combining multi-site diffusion MRI (dMRI) data is hindered by inter-scanner variability, which confounds subsequent analysis. Previous harmonization methods require large, matched or traveling human subjects from multiple sites, which are impractical to acquire in many situations. This study aims to develop a deep learning-based dMRI harmonization framework that eliminates the reliance on multi-site in-vivo traveling human data for training.
\section{Methods} HARP employs a voxel-wise 1D neural network trained on an easily transportable diffusion phantom. The model learns relationships between spherical harmonics coefficients of different sites without memorizing spatial structures.
\section{Results} HARP reduced inter-scanner variability levels significantly in various measures. Quantitatively, it decreased inter-scanner variability as measured by standard error in FA (12\%), MD (10\%), and GFA (30\%) with scan-rescan standard error as the baseline, while preserving fiber orientations and tractography after harmonization.
\section{Conclusion} We believe that HARP represents an important first step toward dMRI harmonization using only phantom data, thereby obviating the need for complex, matched in vivo multi-site cohorts. This phantom-only strategy substantially enhances the feasibility and scalability of quantitative dMRI for large-scale clinical studies.
}


\maketitle

\section{Introduction}\label{introduction}

Diffusion Magnetic Resonance Imaging (dMRI) \cite{johansen2013diffusion} has emerged as a pivotal modality in neuroimaging, providing a unique, non-invasive probe into the microstructural organization of biological tissues \cite{charles2006diffusion, mori2006principles}. By offering quantitative metrics that reflect tissue integrity, dMRI has extended its utility from fundamental neuroscience research to promising clinical applications for assessing the central nervous system. This transition is increasingly catalyzed by the rise of large-scale, multi-center cohort studies, where dMRI plays a central role in the search for robust imaging biomarkers and the longitudinal characterization of neurological conditions.

Despite its clinical and scientific potential, dMRI data exhibit significant systematic variations when acquired across different hardware vendors, magnetic field strengths, or imaging protocols \cite{cetin2018harmonizing, liu2025effect, liu2024reduced, ning2020cross, tax2019cross}. These scanner-specific factors often lead to substantial discrepancies in fundamental image characteristics, such as contrast and signal-to-noise ratio (SNR), even when the same subject is scanned \cite{cetin2024harmonized, schilling2021fiber}. When such multi-site data are pooled for joint analysis, these hardware-driven fluctuations manifest as batch effects, which are non-biological variances that can severely confound the biological signals of interest \cite{warrington2023resource}. Particularly, if the variance attributed to the imaging environment is comparable to or larger than the effect size of the clinical condition, it becomes difficult to distinguish genuine pathological changes from technical artifacts. Consequently, robust data harmonization is an essential prerequisite to decouple these site-specific technical signatures from true biological variability, ensuring the reliability and generalizability of multi-site dMRI findings \cite{fortin2017harmonization, johnson2007adjusting, mirzaalian2016inter, karayumak2019retrospective, nath2019inter, moyer2020scanner, pomponio2020harmonization, descoteaux2024harmonization}.

Existing strategies for dMRI harmonization can be broadly classified into statistical adjustments, signal-based transformations, and deep learning architectures \cite{tax2025multicenter}. Statistical approaches, such as ComBat and its derivatives, primarily target cross-site variances in derived microstructural metrics, including fractional anisotropy (FA) and mean diffusivity (MD) \cite{fortin2017harmonization, johnson2007adjusting, o2011introduction}. In contrast, signal-level methods utilizing spherical harmonics provide a more direct approach to data integration. For instance, techniques based on rotationally invariant spherical harmonics (RISH) features aim to harmonize the diffusion signal itself by calculating voxel-wise scaling factors between source and target sites within the spherical harmonic domain \cite{mirzaalian2016inter, karayumak2019retrospective,de2022cross}. More recently, deep learning-based frameworks have been introduced to capture complex, non-linear mappings between disparate imaging sites, thereby offering enhanced flexibility in modeling site-specific variations compared to traditional linear models \cite{nath2019inter, moyer2020scanner, huynh2019multi}.

Despite the progress in mitigating batch effects, conventional dMRI harmonization strategies are often constrained by stringent requirements for specialized datasets during model construction. A prominent approach involves the use of traveling subjects, where the same individuals are scanned across multiple sites to establish a ground truth for cross-site differences \cite{mirzaalian2016inter, dewey2019deepharmony}. While effective, the recruitment of such subjects imposes significant logistical burdens, including prohibitive costs, extended timelines, and complex ethical considerations. To circumvent the need for traveling subjects, alternative methods such as LinearRISH have been proposed \cite{karayumak2019retrospective, moyer2020scanner, liu2024learning}, which utilize a multi-site dataset with matched distribution at the group level. However, these approaches still necessitate large and meticulously matched cohorts across all participating sites, requiring that demographic variables such as age and sex be equivalently distributed to prevent confounding the harmonization process. The acquisition of either traveling subject data or such matched multi-site cohorts remains highly impractical for most large-scale applications, thereby limiting the scalability and flexibility of existing harmonization frameworks \cite{de2025cross}.

To alleviate the dependency on matched cohorts, recent investigations have explored more flexible harmonization architectures. Some approaches utilize a deep learning network as a prior for a specific site, allowing a model trained on target site data to be applied across diverse source sites \cite{jeong2023blindharmony,  beizaee2025harmonizing}. While these methods show promise, they have primarily been validated on T1-weighted structural images and often fail to explicitly characterize the intricate signal relationships between disparate sites. Another emerging strategy involves accounting for potential demographic biases when learning harmonization parameters using unmatched training groups \cite{de2025cross}. Although this approach demonstrates robust performance, it remains reliant on large-scale in-vivo datasets to achieve statistical stability and may not account for all sources of acquisition bias.

\begin{figure*}
\centering

\includegraphics[width=0.97\textwidth]{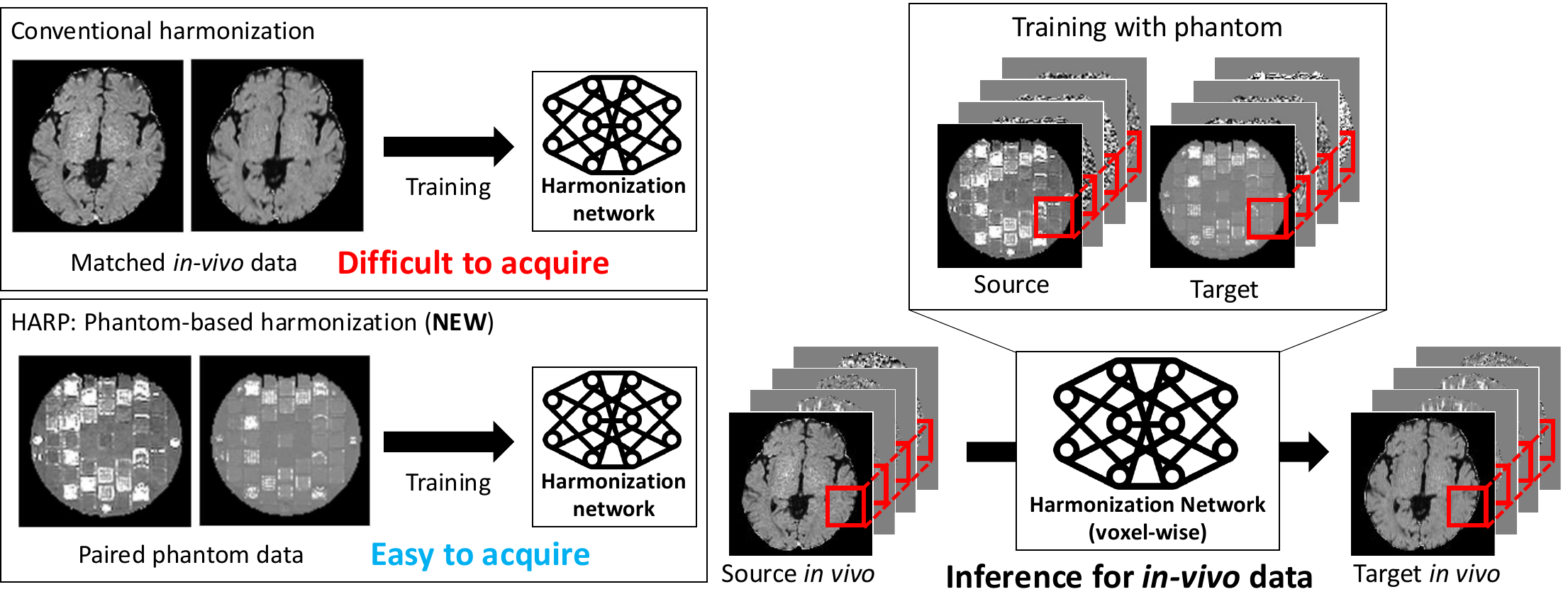}

\caption{Conceptual overview of the HARP framework. (a) Comparison between conventional harmonization and the proposed HARP. Unlike traditional methods that necessitate the acquisition of matched or traveling human subject data from multiple sites, HARP utilizes easily accessible diffusion phantom data for network training. (b) Architectural overview of the 1D voxel-wise neural network. The network receives spherical harmonics coefficients from the source site as input and estimates harmonized spherical harmonics coefficients. This non-spatial architecture ensures robust generalizability when transferring the learned relationships from phantom training to in-vivo human inference.}
\label{fig:concept}
\end{figure*}

Recently, sophisticated diffusion phantoms that closely emulate the microstructural properties of in-vivo white matter tissues have been developed \cite{lavdas2013phantom,wilde2018assessment,aylward2022calibration}. These phantoms offer distinct advantages over human subjects: they are inherently motion-free, maintain stable material properties, and exhibit high reproducibility regardless of environmental fluctuations. Such stability facilitates the efficient and repetitive acquisition of large-scale datasets across diverse scanner hardware and parameter configurations. Furthermore, the ability to manufacture multiple phantoms with identical specifications provides an ideal, controlled environment to establish a uniform reference for training robust harmonization models. If a harmonization framework can be constructed using paired phantom data and subsequently generalized to in-vivo datasets, the logistical and ethical burdens associated with large-scale data acquisition could be substantially mitigated.

Therefore, we propose HARP, a novel deep learning-based framework that utilizes exclusively phantom data for training, yet can be directly applied to in-vivo dMRI data (Fig. \ref{fig:concept}a). The primary objective of this study is to demonstrate that a model optimized on standardized phantoms can successfully capture and correct for site-specific variations, effectively generalizing to the complex architecture of human brain data without the need for matched in-vivo cohorts.

\begin{table*}
\setlength{\tabcolsep}{0.5pt}
\centering
   \caption{Diffusion phantom acquisition protocols for multi-site harmonization. The experimental configurations are illustrated for two distinct scenarios: (a) inter-vendor harmonization between a GE Premier (source site) and a Siemens Prisma (target site), and (b) inter-platform harmonization between a Siemens Skyra (source site) and a Siemens Prisma (target site).}
      {\tiny
   
     \begin{tabular}{ccccc}
   \multicolumn{5}{c}{(a) GE $\rightarrow$ Siemens } \\
   \hline
     $b$ value (s/mm$^2$) & Number of gradients & TE (ms) & TR (ms) &Slice thickness (mm)  \\
    
    \hline
    1000 &	60 &	59.1 &	7500 &	1.5  \\
    1500 &	60 &	63.4 &	9200 &	1.5  \\
    2000 &	60 &	72.4 &	10400 &	1.5  \\
    1000 &	30 &	59.1 &	7000 &	1.5  \\
    2000 &	30 &	72.4 &	12000 &	1.5  \\
    1500 &	30 &	62.2 &	6000 &	3  \\
    2000 &	30 &	71.3 &	6000 &	3  \\
    1500 &	30 &	62.9 &	6500 &	2  \\
    2000 &	30 &	71.9 &	7000 &	2  \\
    1500 &	30 &	63.9 &	10000 &	2.5  \\
    2000 &	30 &	72.9 &	15000 &	2.5  \\
  \hline
   \end{tabular}
   \begin{tabular}{ccccccc}
   \multicolumn{7}{c}{(b) Skyra $\rightarrow$ Prisma } \\
   \hline
     \multirow{2}{*}{$b$ value (s/mm$^2$)} & \multirow{2}{*}{Number of gradients} &   \multicolumn{2}{c}{TE (ms)}  & \multicolumn{2}{c}{TR (ms)} & \multirow{2}{*}{Slice thickness (mm)} \\
     & & Skyra &Prisma & Skyra & Prisma & \\
    \hline
    1000 &	64 &	75 & 54 &	5300 &	3700 &	1.8  \\
    1500 &	64 &	86 & 58 &	6000 &	4300 &	1.8  \\
    2000 &	64 &	96 & 61 &	5800 &	4900 &	1.8  \\
    1000 &	30 &	73 & 54 &	5600 &	3900 &	1.8  \\
    2000 &	30 &	95 & 61 &	5600 &	4400 &	1.8  \\
    3000 &	30 &	109 & 67 &	6200 &	5400 &	1.8  \\ 
    1500 &	30 &	85 & 57 &	3400 &	2400 &	3  \\
    2000 &	30 &	94 & 60 &	3500 &	2700 &	3  \\
    1500 &	30 &	85 & 57 &	5200 &	3600 &	2  \\
    2000 &	30 &	94 & 61 &	5500 &	3900 &	2  \\
    1500 &	30 &	85 & 57 &	4100 &	3000 &	2.5  \\
    2000 &	30 &	94 & 60 &	4400 &	3400 &	2.5  \\
  \hline
   \end{tabular}

   }
\label{table:dataset}
\end{table*}
\section{Methods}\label{methods}

\subsection{HARP} 
The HARP framework aims to train a deep learning model using only phantom data and to apply it to the in-vivo dataset. Most deep learning-based harmonization frameworks employ two-dimensional convolutional neural networks to transform source images into target-like representations by leveraging the ability of convolutional kernels to capture complex spatial patterns \cite{dewey2019deepharmony, liu2021style}. However, when such architectures are trained exclusively on phantom data and subsequently applied to in-vivo datasets, they often exhibit performance degradation due to structural domain shifts. Particularly, the network may overfit to the distinct geometric features of the phantom, which differ fundamentally from the intricate anatomical structures of the human brain \cite{knoll2019assessment, bollmann2019deepqsm}.

To ensure robust generalizability and prevent the network from overfitting to phantom-specific spatial features, we designed HARP as a voxel-wise harmonization framework (Fig. \ref{fig:concept}b). Instead of utilizing spatial kernels, our approach focuses on learning the intrinsic relationship between the spherical harmonics coefficients of the source and target sites within individual voxels.

The core architecture consists of a seven-layer 1D artificial neural network. The network takes a vector of spherical harmonics (SH) coefficients from the source site as input, where the vector length is determined by the maximum order of the SH expansion. Let $C_{ij}^{src}$ denote the spherical harmonics coefficient of order $i$ and degree $j$ from the source site. The network is designed to estimate a set of scaling factors $s_i$, corresponding to each order $i$ to accommodate the unique physical characteristics of the diffusion signal at different angular frequencies. The final harmonized output $C_{ij}^{har}$ is generated by performing an element-wise multiplication between these predicted scales and the original input coefficients:
\begin{equation}
C_{ij}^{har} = s_i \cdot C_{ij}^{src}. 
\label{eq:model}
\end{equation}
The model is optimized using a loss function that minimizes the discrepancy between the harmonized source coefficients $C_{ij}^{har}$ and the corresponding target site spherical harmonics coefficients.

\subsection{Data acquisition}
The performance of HARP was evaluated across two distinct harmonization scenarios. The first scenario involved inter-vendor harmonization between a GE SIGNA Premier (source site) and a Siemens MAGNETOM Prisma (target site). The second scenario focused on intra-vendor, inter-platform harmonization between a Siemens MAGNETOM Skyra (source site) and a Siemens MAGNETOM Prisma (target site). To construct the training datasets, the same diffusion phantoms were scanned at both sites with efforts to maintain consistent phantom positioning and orientation. For validation purposes, in-vivo datasets were acquired from the same subjects at both source and target sites to serve as a traveling subject reference.

For the inter-vendor evaluation, three diffusion phantoms (Phantom 1 \cite{wilde2018assessment,pst}, Phantom 2 \cite{wilde2018assessment,pst}, and Phantom 3 \cite{nist,aylward2022calibration}) were imaged using a 32-channel head coil on the Siemens Prisma and a 48-channel head coil on the GE Premier. Both systems possess identical maximum gradient strengths of 80 mT/m and a maximum slew rate of 200 T/m/s. Vendor-provided simultaneous multi-slice (SMS) echo-planar imaging (EPI) sequences were employed with matched diffusion gradient directions and matched acquisition parameters: 1.5 $\times$ 1.5 mm$^2$ in-plane resolution, 220 $\times$ 220 mm$^2$ field-of-view (FOV), a partial Fourier factor of 6/8, an in-plane acceleration factor of 2, and a multi-band factor of 2, together with $b = 0$ images with AP and PA phase-encoding for EPI distortion correction. Phantom 1 was scanned with a single b-value of 1000 s/mm$^2$, 60 diffusion-encoding directions, and a 1.5 mm slice thickness. Phantoms 2 and 3 were scanned across eleven different parameter configurations (Table \ref{table:dataset}a), varying the b-values, slice thicknesses, and number of diffusion directions. The resulting training library comprised 1,978 slices. For the in-vivo validation (IRB approved), five healthy volunteers participated in same-day scan-rescan sessions using the same protocol as Phantom 1.

For the intra-vendor evaluation, Phantom 2 \cite{wilde2018assessment,pst} and Phantom 3 \cite{nist,aylward2022calibration} were scanned on the Siemens Prisma and the Siemens Skyra, both utilizing 32-channel head coils. These systems differ in their maximum gradient strengths (Prisma: 80 mT/m, Skyra: 45 mT/m) while sharing a maximum slew rate of 200 T/m/s. SMS-EPI sequences were used with matched parameters except for echo time (TE) and repetition time (TR) due to hardware constraints: 1.8 $\times$ 1.8 mm$^2$ in-plane resolution, 220 $\times$ 220 mm$^2$ FOV, 6/8 partial Fourier, in-plane acceleration of 2, and multi-band factor of 2, together with $b = 0$ images with AP and PA phase-encoding for EPI distortion correction. The phantom protocols varied b-values, slice thicknesses, and diffusion directions (Table \ref{table:dataset}b), yielding a total of 1,680 slices. The in-vivo validation cohort (IRB approved) included five healthy subjects scanned with two distinct protocols: (1) $b = 1000$ s/mm$^2$, 64 directions, 1.8 mm slice thickness, with TE/TR of 75/5300 ms on the Skyra and 54/3700 ms on the Prisma; and (2) $b = 2000$ s/mm$^2$, 64 directions, 1.8 mm slice thickness, with TE/TR of 96/5800 ms on the Skyra and 61/4900 ms on the Prisma.

\subsection{Data processing}
The DICOM images for both phantom and in-vivo datasets underwent a comprehensive preprocessing pipeline. To mitigate geometric distortions and motion-induced artifacts, susceptibility-induced off-resonance fields were estimated and corrected using the top-up tool \cite{andersson2003correct}, followed by eddy current and motion correction \cite{andersson2016integrated}. In the case of the Siemens Skyra data, an additional registration step using FSL FLIRT \cite{jenkinson2002improved} was performed to further refine the eddy current correction. Following the correction of the diffusion-weighted images, spherical harmonics coefficients were estimated up to a maximum order of 8 using the DIPY software library \cite{karayumak2019retrospective, mirzaalian2016inter}. For the in-vivo datasets, brain masks were generated using FSL BET \cite{smith2002fast}. To address minor spatial misalignments between the source and target site scans for the phantom datasets, a rigid-body registration was performed using FSL FLIRT. Particularly, the rotation matrix was estimated based on the non-diffusion-weighted ($b = 0$) images. This estimated transformation was subsequently applied to the spherical harmonics coefficient maps.

\subsection{Training details} 
The HARP framework was implemented using a 1D artificial neural network. The input layer consists of 45 neurons, corresponding to the spherical harmonics coefficients up to a maximum order of 8. To estimate the order-specific scaling factors, the output layer was designed with 5 neurons, each representing a scale $s_i$ for orders $i = 0, 2, 4, 6, 8$. The architecture comprises seven hidden layers with 160, 240, 320, 360, 480, 520, and 600 fully connected neurons, respectively. To facilitate gradient flow and preserve feature integrity, skip connections were integrated by embedding the input data into the third, fifth, and seventh hidden layers. Rectified Linear Unit (ReLU) activation functions were employed for the hidden layers. To ensure high-quality training, the phantom dataset was pre-filtered to exclude noise and nuisance voxels. Particularly, voxels were discarded if their zeroth-order spherical harmonics coefficient value was lower than the 30th percentile of the corresponding coefficient in the in-vivo source site images. The network was trained only with the phantom data for 40,000 iterations using an $L_2$ loss function, a learning rate of $10^{-6}$, and the Adam optimizer. During the inference stage, the spherical harmonics coefficients of the in-vivo source site data were fed into the trained network to generate the harmonized spherical harmonics coefficients through the estimated order-specific scales.

\begin{figure*}
\centering
\includegraphics[width=0.95\textwidth]{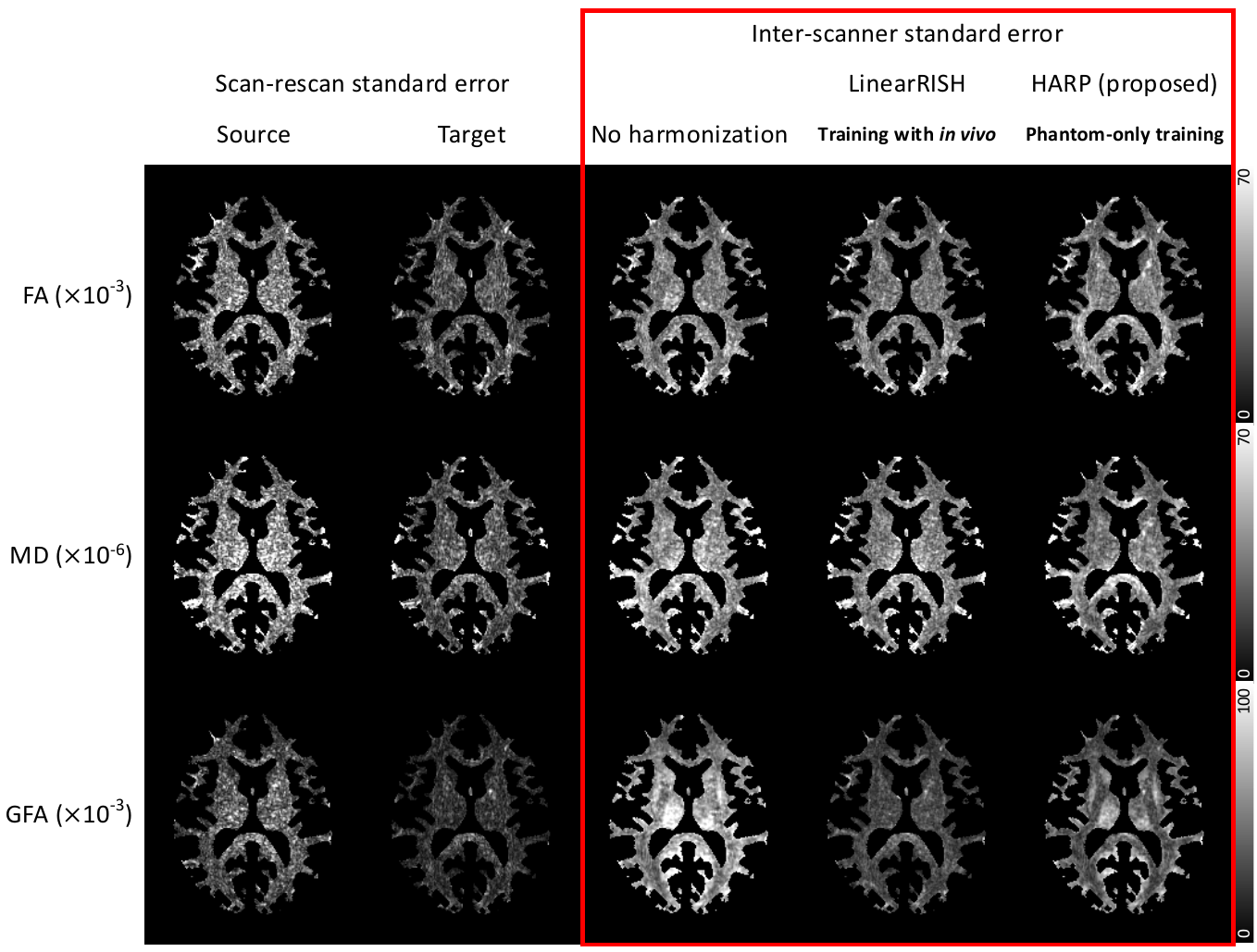}

\caption{Voxel-wise standard error maps for GE $\rightarrow$ Siemens harmonization. The spatial distribution of standard errors for fractional anisotropy (FA), mean diffusivity (MD), and generalized fractional anisotropy (GFA) is displayed. Columns 1 and 2 represent the scan-rescan standard error for the source and target sites, respectively. Prior to harmonization (Column 3), the inter-scanner standard error is notably high across all metrics. The application of HARP (Column 5), which is trained exclusively on phantom data, results in a substantial reduction of standard error, achieving performance closer to the in-vivo-trained LinearRISH baseline (Column 4).}
\label{fig:voxelSE}
\end{figure*}

\begin{table*}
\centering

   \caption{Average voxel-wise standard error values for FA, MD, and GFA across three harmonization scenarios. The inter-scanner standard error is reported for the No harmonization data, LinearRISH, and the proposed HARP. Values in parentheses represent the percentage difference between inter-scanner standard error and source site scan-rescan standard error: ($\frac{\sigma_{e,Inter-scanner}-\sigma_{e,Scan-rescan}}{\sigma_{e,Scan-rescan}}$).} 
   
      {\tiny
     \begin{tabular}{cc|cc|ccc}
   \hline
     & & \multicolumn{2}{c}{Scan-rescan standard error}  &  \multicolumn{3}{c}{Inter-scanner standard error} \\
    & & Source & Target & No harmonization & LinearRISH & HARP \\
    \hline
    \multirow{3}{*}{GE $\rightarrow$ Siemens} &FA ($\times10^{-3}$)& 42.2 $\pm$ 24.5 & 26.6 $\pm$ 15.9 & 46.5 (10\%) $\pm$ 23.1 & 39.9 (-5\%) $\pm$ 19.1 & 41.6 (-1\%) $\pm$ 18.4 \\
    &MD ($\times10^{-6}$)& 60.6 $\pm$ 60.9 & 36.9 $\pm$ 36.5  & 65.7 (8\%) $\pm$ 63.2 & 59.3 (-2\%) $\pm$ 57.3 & 55.2 (-9\%) $\pm$ 55.1 \\
    &GFA ($\times10^{-3}$)& 42.7 $\pm$ 26.5 & 23.5 $\pm$ 16.1  & 81.9 (92\%) $\pm$ 37.9 & 34.4 (-19\%) $\pm$ 20.0 & 56.9 (33\%) $\pm$ 28.6 \\
  \hline
    \multirow{2}{*}{Skyra $\rightarrow$ Prisma } & FA ($\times10^{-3}$)& 28.4 $\pm$ 23.8 & 25.1 $\pm$ 14.0 & 44.4 (56\%) $\pm$ 30.2 & 37.8 (33\%) $\pm$ 20.3 & 40.3 (42\%) $\pm$ 24.0 \\
    &MD ($\times10^{-6}$)& 48.0 $\pm$ 56.1 & 48.0 $\pm$ 48.4  & 80.7 (68\%) $\pm$ 93.8 & 74.9 (56\%) $\pm$ 93.1 & 77.0 (61\%) $\pm$ 92.7 \\
    $b = 1000 $ s/mm$^2$ &GFA ($\times10^{-3}$)& 25.9 $\pm$ 24.1 & 22.0 $\pm$ 14.2  & 38.9 (50\%) $\pm$ 29.1 & 32.5 (25\%) $\pm$ 19.4 & 34.9 (35\%) $\pm$ 23.6 \\
  \hline
    \multirow{2}{*}{Skyra $\rightarrow$ Prisma} &FA ($\times10^{-3}$)& 27.5 $\pm$ 19.6 & 21.7 $\pm$ 11.5 & 35.0 (27\%) $\pm$ 20.8 & 32.3 (17\%) $\pm$ 17.0 & 32.2 (17\%) $\pm$ 17.0 \\
    &MD ($\times10^{-6}$)& 33.5 $\pm$ 33.2 & 25.6 $\pm$ 24.4  & 46.2 (38\%) $\pm$ 48.6 & 43.1 (29\%) $\pm$ 49.1 & 43.7 (31\%) $\pm$ 49.1 \\
    $b = 2000 $ s/mm$^2$ &GFA ($\times10^{-3}$)& 24.9 $\pm$ 20.0 & 18.4 $\pm$ 10.7  & 34.6 (39\%) $\pm$ 22.6 & 27.7 (11\%) $\pm$ 16.4 & 30.3 (22\%) $\pm$ 19.0 \\
  \hline
   \end{tabular}
   }
\label{table:voxelSE}
\end{table*}

\subsection{Competing method}
LinearRISH \cite{karayumak2019retrospective}, a retrospective harmonization method that requires in-vivo data from traveling subjects for model construction, served as the competing method for performance evaluation. Compared to phantom-only training in HARP, LinearRISH was trained and evaluated using the same in-vivo traveling subject dataset in this study, so it represents a best-case scenario for harmonization performance when matched human references are available.

\subsection{Experiments} 
To quantify the scan-rescan reproducibility and inter-scanner variability, we utilized the standard error ($\sigma_e$) \cite{liu2025effect, matheson2019we}, defined as:
\begin{equation}
\sigma_e^2 = \frac{1}{N(K-1)} \sum_{i=1}^{N}\sum_{j=1}^{K}(y_{ij}-\bar{y_i})^2, 
\label{eq:se}
\end{equation}
where $N$ is the number of subjects, $i$ is the subject index, $j$ is the measurement index, and $K$ denotes the number of measurements per subject. Particularly, $K = 2$ was assigned for scan-rescan variability assessments, while $K = 4$ was used for inter-scanner comparisons. In this context, $y_{ij}$ represents the dMRI-derived metric, and $\bar{y_i}$ is the mean value of the measurements for a given subject.

The harmonized spherical harmonics coefficients were used to reconstruct diffusion-weighted images based on the acquired diffusion gradient tables. From these reconstructed signals, fractional anisotropy (FA) and mean diffusivity (MD) maps were estimated using the diffusion tensor model \cite{o2011introduction}, and generalized fractional anisotropy (GFA) was calculated from the orientation distribution function (ODF) estimated via the constrained spherical deconvolution (CSD) \cite{tournier2007robust}. To facilitate group-wise comparisons, these quantitative maps were registered to the IIT human brain atlas space \cite{zhang2018evaluation} using ANTs non-rigid registration. The transformation fields were estimated from the FA maps and subsequently applied to the MD and GFA maps. Voxel-wise standard errors were computed across the normalized maps to quantify variability, and the mean standard error was calculated within the brain mask, excluding cerebrospinal fluid (CSF) regions. The percentage difference between inter-scanner standard error and source site scan-rescan standard error was also calculated as: ($\frac{\sigma_{e,Inter-scanner}-\sigma_{e,Scan-rescan}}{\sigma_{e,Scan-rescan}}$). Here, the maximum of the source and target scan-rescan standard error was used (for $\sigma_{e,Inter-scanner}$) as that is the lowest inter-scanner variability one can hope to achieve (as it will be similar to scan-rescan variability). Thus, a number closer to zero would indicate that a harmonization method was successful in reducing inter-scanner variability closer to that of scan-rescan variability.

Region-of-interest (ROI) analysis was performed for specific white matter tracts, including the forceps major, forceps minor, cingulum, corticospinal tract, inferior longitudinal fasciculus, inferior fronto-occipital fasciculus, superior longitudinal fasciculus, and uncinate fasciculus. Gray matter regions, including the frontal, parietal, temporal, and occipital lobes, as well as the cerebellum, were also evaluated. All ROI masks were derived from the IIT human brain atlas \cite{zhang2018evaluation} and warped into the native subject space using the inverse of the previously calculated deformation fields. Standard errors were then calculated for the ROI-averaged values, and paired t-tests were conducted for standard errors to assess the statistical significance of the differences between methods.

\begin{figure*}
\centering
\includegraphics[width=0.99\textwidth]{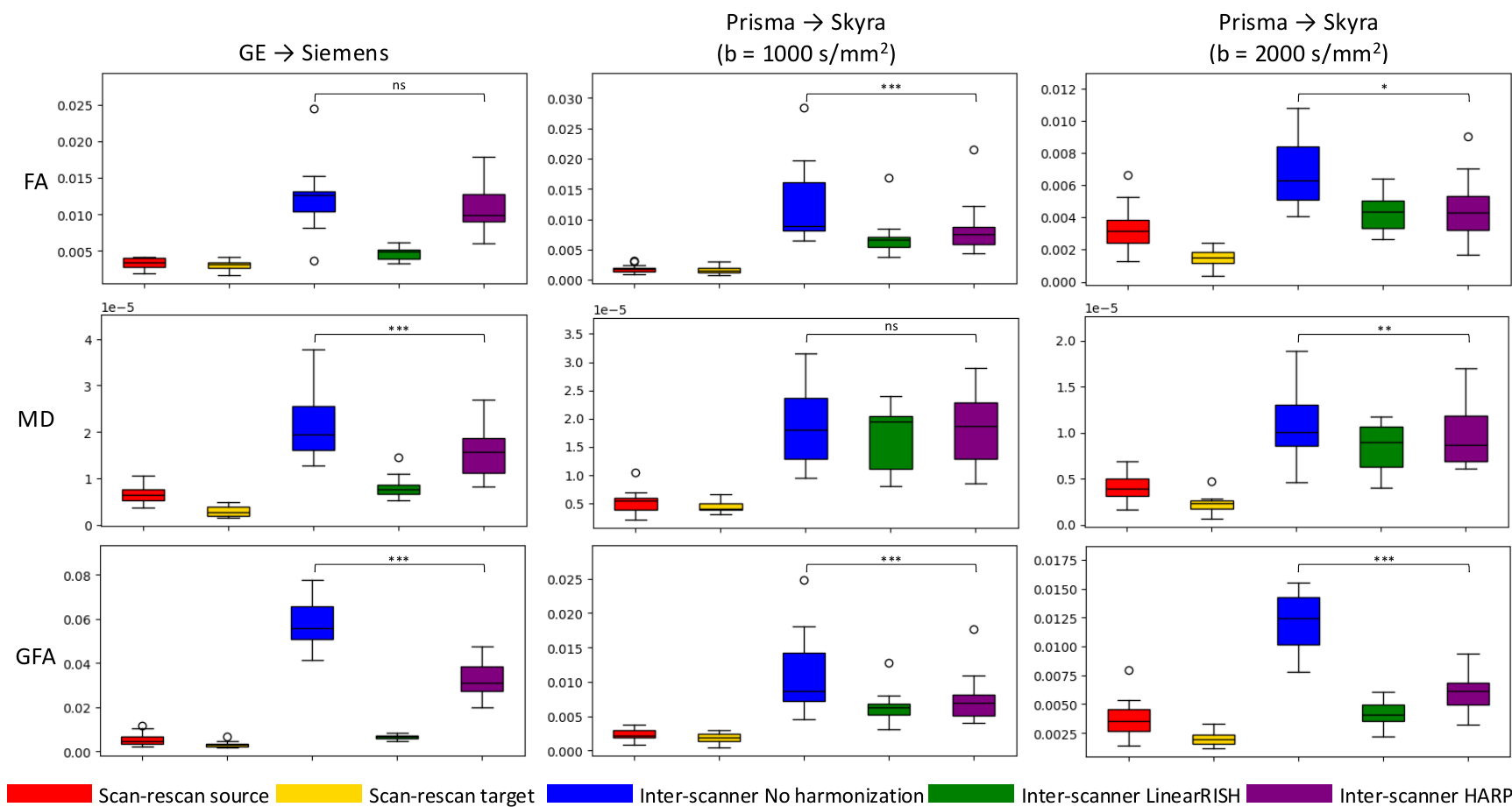}

\caption{Box plots illustrating the standard errors for averaged FA, MD, and GFA values across selected white matter regions of interest. The variability is categorized into source site scan-rescan (red), target site scan-rescan (yellow), inter-scanner without harmonization (blue), inter-scanner with LinearRISH (green), and inter-scanner with HARP (purple). Statistical significance from paired t-tests is indicated by *, **, and *** for p-values less than 0.05, 0.01, and 0.001, respectively. HARP significantly reduces the standard error across most tracts, with the exception of FA in the GE $\rightarrow$ Siemens scenario and MD in the Skyra $\rightarrow$ Prisma scenario at b = 1000 s/mm$^2$.}
\label{fig:roiSE_wm}
\end{figure*}

To further ensure that the harmonization process preserved the integrity of fiber orientation information, we performed CSD-based fiber orientation estimation \cite{tournier2007robust} and UKF tractography using a multi-tensor model \cite{malcolm2009two}. White matter tract parcellation was executed using the whitematteranalysis (WMA) toolbox \cite{o2007automatic}. The accuracy of the fiber orientation was evaluated using the angular error of the principal fiber directions within the white matter mask \cite{schilling2016comparison}. Additionally, tractography consistency was quantified using the weighted DICE similarity (wDICE) \cite{zhang2019test}, comparing the results between the original source site data and the harmonized HARP outputs.

\section{Results}\label{results}
\subsection{Voxel-wise analysis}

Voxel-wise standard error maps for the GE $\rightarrow$ Siemens harmonization scenario are illustrated in Figure \ref{fig:voxelSE}. The first and second columns display the scan-rescan standard error for the source and target sites, respectively, while the subsequent columns show the inter-scanner variability without harmonization (column 3), with LinearRISH (column 4), and with the proposed HARP framework (column 5). The data harmonized by HARP consistently demonstrated lower voxel-wise standard error compared to the No harmonization results. Furthermore, the performance of the phantom-trained HARP was closer to LinearRISH results, which represents the best-case scenario utilizing in-vivo traveling subject data for training. Consistent trends were observed in the Siemens Skyra $\rightarrow$ Siemens Prisma scenario, the results of which are provided in the Supporting Information.

The averaged voxel-wise standard errors within the brain mask (excluding CSF) are quantitatively summarized in Table \ref{table:voxelSE}. This table reports the scan-rescan standard errors for both sites alongside the inter-scanner standard errors for the No harmonization, LinearRISH, and HARP. Compared to the No harmonization case, HARP successfully reduced the percentage difference between inter-scanner standard error and scan-rescan standard error by 12\% for FA, 10\% for MD, and 30\% for GFA. These quantitative improvements are closer to those achieved by LinearRISH, validating that a phantom-trained model can effectively mitigate site-specific biases without the necessity of human training data.

\begin{figure*}
\centering
\includegraphics[width=0.99\textwidth]{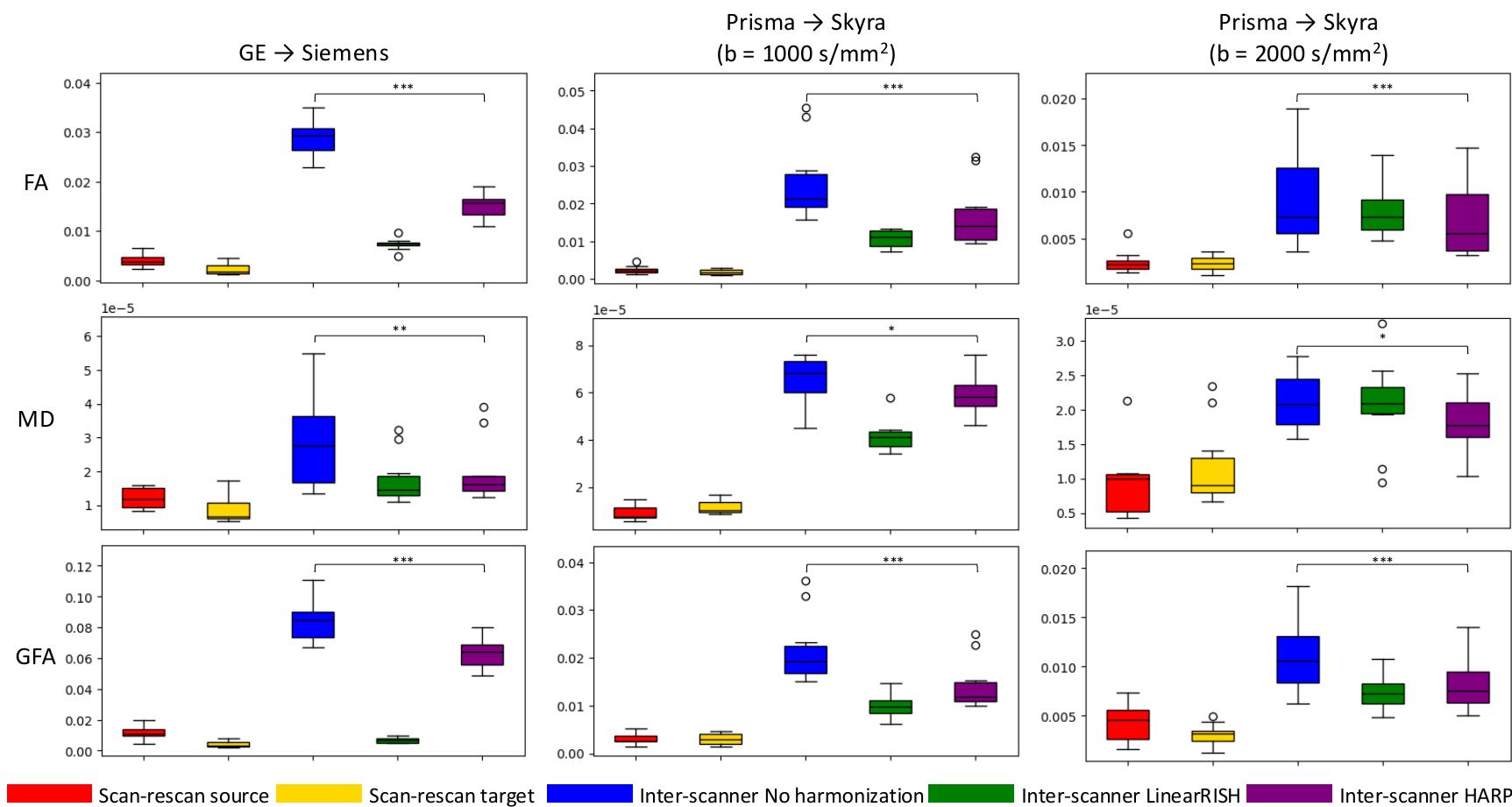}

\caption{Box plots illustrating the standard errors for averaged FA, MD, and GFA values across selected gray matter regions of interest. The color coding follows the same scheme as in Figure \ref{fig:roiSE_wm}: source site scan-rescan (red), target site scan-rescan (yellow), inter-scanner without harmonization (blue), inter-scanner with LinearRISH (green), and inter-scanner with HARP (purple). Significance levels are denoted by *, **, and *** for $p < $ 0.05, 0.01, and 0.001, respectively. In gray matter regions, HARP consistently and significantly reduces the inter-scanner standard error across all cases, demonstrating robust harmonization that aligns with the performance of the in-vivo-trained baseline.}
\label{fig:roiSE_gm}
\end{figure*}

\begin{table*}
\centering
   \caption{Quantitative summary of mean standard error values for FA, MD, and GFA averaged across white matter (WM) and gray matter (GM) regions of interest. }
      {\tiny
     \begin{tabular}{ccc|cc|ccc}
   \hline
     & & & \multicolumn{2}{c}{Scan-rescan standard error}  &  \multicolumn{3}{c}{Inter-scanner standard error} \\
     & & &  Source & Target & No harmonization & LinearRISH & HARP (proposed) \\
\hline
\multirow{6}{*}{GE $\rightarrow$ Siemens} & \multirow{2}{*}{FA ($\times10^{-3}$)} & WM & 3.33 $\pm$ 0.75 & 3.04 $\pm$ 0.76 & 12.2 $\pm$ 4.58 & 4.64 $\pm$ 0.92 & 11.0 $\pm$ 3.27 \\
    & & GM & 4.09 $\pm$ 1.37 & 2.21 $\pm$ 1.13 & 28.8 $\pm$ 3.55 & 7.31 $\pm$ 1.21 & 15.1 $\pm$ 2.56 \\
    &\multirow{2}{*}{MD ($\times10^{-6}$)} & WM & 6.47 $\pm$ 1.77 & 2.90 $\pm$ 1.18 & 22.1 $\pm$ 8.03 & 8.07 $\pm$ 2.38 & 15.9 $\pm$ 5.90 \\
    & & GM & 12.1 $\pm$ 3.07 & 8.95 $\pm$ 4.61 & 29.9 $\pm$ 15.5 & 17.6 $\pm$ 7.44 & 19.7 $\pm$ 9.34 \\
    &\multirow{2}{*}{GFA ($\times10^{-3}$)} & WM & 5.45 $\pm$ 2.87 & 3.16 $\pm$ 1.30 & 58.1 $\pm$ 10.0 & 6.73 $\pm$ 0.96 & 32.5 $\pm$ 8.28 \\
    & & GM & 11.8 $\pm$ 4.10 & 3.96 $\pm$ 2.07 & 84.5 $\pm$ 14.4 & 6.69 $\pm$ 1.59 & 63.2 $\pm$ 9.61 \\
  \hline

\multirow{4}{*}{Skyra $\rightarrow$ Prisma} & \multirow{2}{*}{FA ($\times10^{-3}$)} & WM & 1.78 $\pm$ 0.72 & 1.65 $\pm$ 0.68 & 12.5 $\pm$ 6.54 & 6.98 $\pm$ 3.09 & 8.42 $\pm$ 4.37 \\
    & & GM & 2.32 $\pm$ 1.09 & 1.79 $\pm$ 0.66 & 25.6 $\pm$ 10.5 & 10.7 $\pm$ 2.35 & 16.9 $\pm$ 8.58 \\
    & \multirow{2}{*}{MD ($\times10^{-6}$)} & WM & 5.15 $\pm$ 2.10 & 4.39 $\pm$ 1.01 & 18.9 $\pm$ 7.33 & 16.8 $\pm$ 5.56 & 18.6 $\pm$ 6.88 \\
    & & GM & 9.03 $\pm$ 3.01 & 11.7 $\pm$ 3.19 & 65.8 $\pm$ 10.1 & 41.7 $\pm$ 6.76 & 59.2 $\pm$ 8.98 \\
    \multirow{2}{*}{$b = 1000$ s/mm$^2$} & \multirow{2}{*}{GFA ($\times10^{-3}$)} & WM & 2.33 $\pm$ 0.87 & 1.83 $\pm$ 0.69 & 11.2 $\pm$ 5.89 & 6.27 $\pm$ 2.31 & 7.50 $\pm$ 3.64 \\
    & & GM & 3.12 $\pm$ 1.15 & 3.03 $\pm$ 1.23 & 21.6 $\pm$ 7.28 & 9.96 $\pm$ 2.69 & 14.2 $\pm$ 5.35 \\
    \hline

\multirow{4}{*}{Skyra $\rightarrow$ Prisma} & \multirow{2}{*}{FA ($\times10^{-3}$)} & WM & 3.33 $\pm$ 1.43 & 1.47 $\pm$ 0.59 & 6.90 $\pm$ 2.38 & 4.31 $\pm$ 1.10 & 4.53 $\pm$ 1.90 \\
    & & GM & 2.46 $\pm$ 1.23 & 2.29 $\pm$ 0.86 & 9.26 $\pm$ 5.20 & 7.82 $\pm$ 2.85 & 6.93 $\pm$ 4.06 \\
    & \multirow{2}{*}{MD ($\times10^{-6}$)} & WM & 4.13 $\pm$ 1.52 & 2.28 $\pm$ 0.97 & 11.2 $\pm$ 3.91 & 8.54 $\pm$ 2.64 & 9.49 $\pm$ 3.24 \\
    & & GM & 9.31 $\pm$ 5.03 & 11.7 $\pm$ 5.95 & 21.3 $\pm$ 4.16 & 20.6 $\pm$ 6.61 & 18.2 $\pm$ 4.17 \\
    \multirow{2}{*}{$b = 2000$ s/mm$^2$} & \multirow{2}{*}{GFA ($\times10^{-3}$)} & WM & 3.73 $\pm$ 1.70 & 1.99 $\pm$ 0.56 & 12.2 $\pm$ 2.55 & 4.13 $\pm$ 1.06 & 6.12 $\pm$ 1.75 \\
    & & GM & 4.30 $\pm$ 1.89 & 3.10 $\pm$ 1.08 & 11.0 $\pm$ 3.85 & 7.39 $\pm$ 1.71 & 8.45 $\pm$ 2.87 \\
    \hline
   \end{tabular}
   }
\label{table:roiSE}
\end{table*}

\subsection{ROI study}
The scan-rescan and inter-scanner standard errors for the averaged fractional anisotropy (FA), mean diffusivity (MD), and generalized fractional anisotropy (GFA) values within white matter and gray matter regions are presented in Figure \ref{fig:roiSE_wm} and Figure \ref{fig:roiSE_gm}, respectively. The variability is color-coded to distinguish among the source site scan-rescan (red), target site scan-rescan (yellow), inter-scanner without harmonization (blue), inter-scanner with LinearRISH (green), and inter-scanner with HARP (purple). The HARP framework, despite being trained exclusively on phantom data, effectively reduced batch effects across most regions. The resulting standard error values were closer to those obtained with the in-vivo-trained LinearRISH.

While HARP demonstrated strong overall performance, its efficacy in a few specific white matter cases (FA in the GE $\rightarrow$ Siemens scenario and MD in the Skyra $\rightarrow$ Prisma scenario at $b = 1000$ s/mm$^2$) was slightly lower than in other evaluations. This observation may stem from the inherent limitations of current diffusion phantoms. Although these phantoms are designed to emulate complex microstructures such as crossing fibers in the white matter, they remain simplified models that do not fully capture the intricate anatomical realism of the human brain.

\begin{figure*}
\centering

\includegraphics[width=0.95\textwidth]{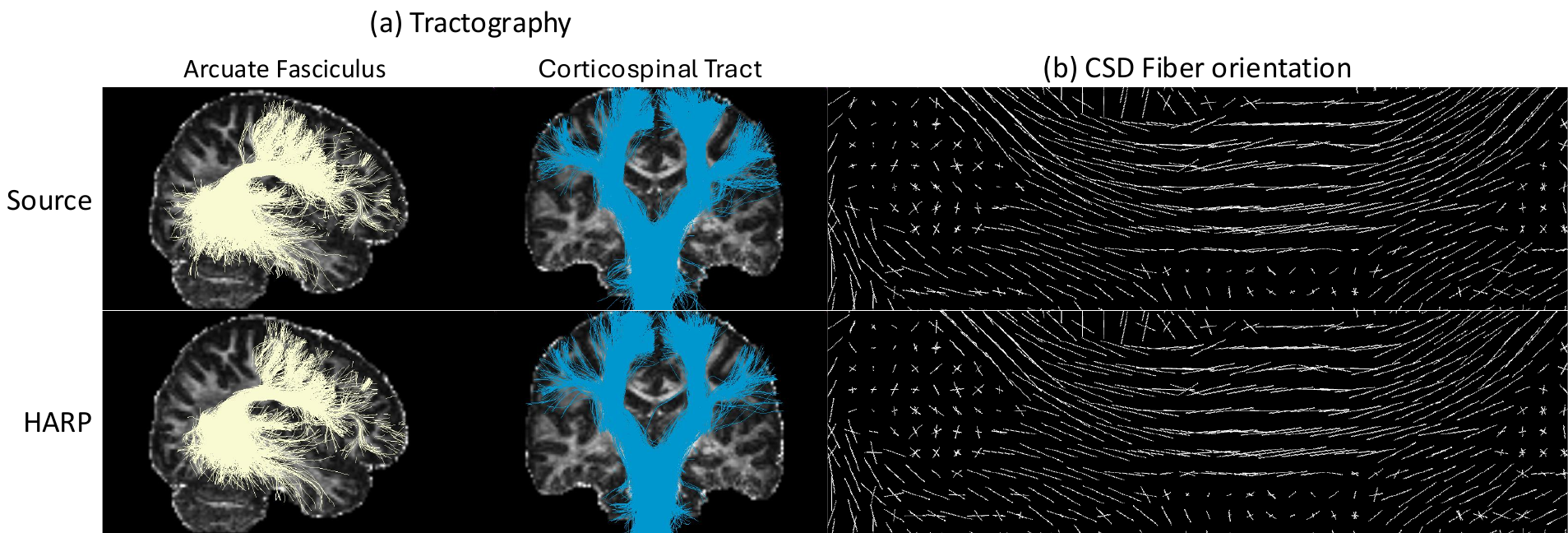}

\caption{Visual comparison of (a) tractography (arcuate fasciculus and corticospinal tract) and (b) fiber orientation maps before (top) and after (bottom) HARP harmonization. The high consistency between rows confirms that the 1D voxel-wise framework preserves underlying fiber information.}
\label{fig:fiber}
\end{figure*}

The averaged standard error values for FA, MD, and GFA across all white matter and gray matter regions are quantitatively detailed in Table \ref{table:roiSE}. The table compares scan-rescan variability with inter-scanner variability across the different harmonization methods. The quantitative results confirm that HARP successfully reduces systematic bias, achieving a degree of harmonization that matches the performance of methods requiring in-vivo traveling subject data.

\subsection{Fiber analysis}
The impact of HARP harmonization on the geometric integrity of the diffusion signal is illustrated in Figure \ref{fig:fiber}. Representative white matter tracts, particularly the arcuate fasciculus and the corticospinal tract, are visualized alongside the CSD fiber orientation maps for the source site data before and after harmonization. The tractography results remained consistent following the application of HARP, with no discernible loss of structural connectivity or anatomical detail. Similarly, the fiber orientation maps demonstrated high spatial stability, confirming that the transformation learned from the phantom data does not introduce artificial rotations or distortions into the spherical harmonics domain.

Quantitatively, the high degree of similarity between the original source data and the harmonized results was confirmed through rigorous geometric metrics. The tractography maps achieved a high weighted DICE similarity (wDICE) score of 0.885, reflecting excellent preservation of the reconstructed fiber bundles. Furthermore, the fiber orientation maps exhibited an angular error of 6.1$^\circ$. This high level of consistency in fiber-related metrics indicates that HARP preserves the essential fiber orientation information while successfully harmonizing the signal characteristics across disparate imaging sites.

\begin{table}
\centering
   \caption{Weighted DICE (wDICE) and angular error values comparing dMRI data before and after HARP harmonization. Across all scenarios, high wDICE scores and low angular errors demonstrate that HARP preserves the underlying fiber information.}
      {\tiny
     \begin{tabular}{c|ccc}
   \hline
     &   \multirow{2}{*}{ GE $\rightarrow$ Siemens } & Skyra $\rightarrow$ Prisma & Skyra $\rightarrow$ Prisma \\
     &&  $b = 1000 $ s/mm$^2$ & $b = 2000 $ s/mm$^2$ \\
    \hline
    wDICE& 0.872 & 0.874 & 0.910 \\
    Angular error& 6.49$^\circ$ & 6.83$^\circ$ & 4.93$^\circ$ \\
    \hline
   \end{tabular}
   }

\label{table:fiber}
\end{table}

\section{Discussion}\label{discussion}
In this study, we introduced HARP, a one-dimensional voxel-wise neural network-based harmonization framework designed to be trained exclusively on phantom data and directly applied to in-vivo human dMRI datasets. Unlike conventional dMRI harmonization strategies that necessitate traveling subjects or matched cohorts, the proposed HARP framework eliminates the logistical and ethical complexities associated with large-scale in-vivo data acquisition. Our results demonstrate that HARP achieves performance closer to LinearRISH, which represents an idealized scenario utilizing matched human references. 

The traditional process of building multi-site in-vivo datasets (e.g., comprising subject recruitment, rigorous screening, and prolonged MRI sessions) is inherently burdensome and time-consuming. In contrast, diffusion phantoms offer superior clinical utility due to their portability and the absence of participant-related constraints. Phantoms allow for the rapid acquisition of extensive datasets across diverse parameter configurations without the risk of patient discomfort or motion artifacts. Furthermore, if multiple phantoms can be manufactured with identical microstructural specifications, the requirement for a single traveling phantom is removed. Researchers can instead scan separate but identical phantoms at disparate sites to train a unified deep learning model, significantly enhancing the scalability of multi-center harmonization efforts.

The adoption of a 1D architecture is a critical design choice to prevent the network from memorizing the specific geometric structures of the phantom. By processing data at the voxel level, the model utilizes inter-dimensional information within the spherical harmonics coefficients and leverages inductive biases to model complex signal relationships more accurately than traditional linear methods (See Supporting Information). Most existing deep learning-based harmonization frameworks, which typically rely on 2D or 3D convolutional neural networks, suffer from limited generalizability when applied to pathological data. These spatial architectures tend to encode the specific anatomical structures of the training set, leading to degraded performance on unseen morphologies such as lesions or tumors \cite{jung2020exploring}. Because the 1D framework of HARP is inherently blind to spatial anatomy, it offers superior generalizability to diverse and potentially pathological in-vivo brain structures.

Despite these advantages, HARP is subject to certain limitations. First, the performance of the model is heavily dependent on the quality and diversity of the training phantoms. Our results indicate that current phantoms lack sufficient complexity to fully emulate the intricate white matter fiber crossings found in the human brain, which may constrain the precision of white matter harmonization. Furthermore, few phantoms have diffusivity and FA values similar to those of gray matter in the human brain, making it difficult for the algorithm to learn the large variability of the tissue types across the brain. Although we incorporated diverse scan parameters, the use of only three phantoms suggests that further improvements will require the development and inclusion of more anatomically realistic (in terms of diffusivities or white matter crossings, etc.) phantoms.

Second, the 1D nature of the network precludes the inclusion of spatial context, as it focuses solely on voxel-to-voxel relationships between the source and target sites. While this approach is effective for capturing global systematic variations, its performance may be limited in cases where signal discrepancies exhibit high spatial frequency variations. Future studies could address this by incorporating limited in-vivo datasets into the training process to refine spatially varying parameters. Furthermore, the hyperparameter configurations in this study are optimized based on the specific phantoms utilized \cite{oh2025fair}. When applying HARP to different phantom types or quantities, re-tuning these parameters against a small in-vivo validation set is recommended to ensure optimal results for each specific study.

\section{Conclusion}\label{conclusion}
We proposed HARP, a deep learning-based harmonization framework trained exclusively on phantom data and directly applicable to in-vivo dMRI. By employing a 1D voxel-wise architecture, the model effectively learned site-to-site relationships from phantoms and successfully generalized to human brain data. Across three distinct in-vivo scenarios, HARP demonstrated performance closer to methods trained on traveling human subjects. This approach significantly reduces the logistical burden of acquiring matched in-vivo datasets and enhances the practical feasibility of multi-scanner dMRI harmonization in clinical research.

\subsection*{Phantom data availability statement}
The phantom and in-vivo data used for this study can be shared upon request. The code for the proposed method is available at \url{https://github.com/Hwihuni/HARP}.

\subsection*{Conflict of interest}

The authors declare no potential conflict of interests. \textbf{NIST Disclaimer}: Certain commercial equipment, instruments, software, or materials are identified in this paper in order to specify the experimental procedure adequately. Such identification is not intended to imply recommendation of endorsement by NIST, nor is it intended to imply that the materials or equipment identified are necessarily the best available for the purpose.

\bibliography{References}%
\hbox{}

\noindent \textbf{Supporting Information}

\noindent Additional supporting information can be found online in the Supporting Information section. \textbf{Figure S1} Voxel-wise standard error maps for Skyra $\rightarrow$ Prisma harmonization in $b = 1000$ s/mm$^2$. \textbf{Figure S2} Voxel-wise standard error maps for Skyra $\rightarrow$ Prisma harmonization in $b = 2000$ s/mm$^2$. \textbf{Figure S3} Comparison results between the linear model (column 5) and HARP (column 6) \textbf{Table S1, S2, S3} Quantitative summary of standard error values for FA, MD, and GFA for each ROI

\end{document}